\documentclass[conference]{IEEEtran}
\IEEEoverridecommandlockouts
\usepackage{cite}
\usepackage{amsmath,amssymb,amsfonts}
\usepackage{algorithmic}
\usepackage{graphicx}
\usepackage{textcomp}
\usepackage{xcolor}
\usepackage{booktabs}
\graphicspath{{figures/}}

\def\BibTeX{{\rm B\kern-.05em{\sc i\kern-.025em b}\kern-.08em
    T\kern-.1667em\lower.7ex\hbox{E}\kern-.125emX}}

\begin{document}

\title{A Lightweight Hybrid Transformer-CRF Architecture for Multi-Type Bangla Medical Entity Recognition}

\author{
\IEEEauthorblockN{Peyal Saha}
\IEEEauthorblockA{
Dept. of CSE\\
KUET, Khulna, Bangladesh\\
peyalsaha455@gmail.com
}
\and
\IEEEauthorblockN{Ahsanul Haque Hasib}
\IEEEauthorblockA{
Dept. of CSE\\
KUET, Khulna, Bangladesh\\
ahsanulhasib2@gmail.com
}
\and
\IEEEauthorblockN{Shoumik Barman Polok}
\IEEEauthorblockA{
Dept. of CSE\\
KUET, Khulna, Bangladesh\\
polokbarman874@gmail.com
}
}

\maketitle

\begin{abstract}
MedER refers to the identification of medical entities. It is crucial for structured clinical information extraction from unstructured data. Many of the systems rely on transformer-based models, which are computationally intensive and would be hard to deploy. Furthermore, earlier works utilize relaxed evaluation metrics which distort performance artificially as they reward predicting “Outside” (O) tokens correctly in the background. In this paper, we propose a lightweight, MedER framework for the Bangla language. We establish a rigorous baseline using a 12-layer BanglaBERT model combined with a Conditional Random Field (CRF) layer for exact-boundary entity detection. To address deployment constraints, we compress this teacher into a 4-layer student network via Knowledge Distillation (KD). Here the student learns from the pre-CRF soft emission logits of the teacher. We then apply INT8 dynamic quantization to further reduce the model size. Our final quantized student runs 8.6× faster on a standard CPU and requires nearly 48\% less storage than the CRF teacher.
\end{abstract}


\section{Introduction}
The use of NLP in the medical field is going to change how data is processed and fed.  MedER is a subTask of NLP (Natural Language Processing) which aims to automatically extract clinical term of medical interest i.e. drug name,disease and organ name from clinical text\cite{b1}.   Recently, significant progress has been achieved on MedER in languages with high resources like English with support from domain rooted models like ClinicalBERT and BioBERT   \cite{b2,b3}. However, it is used on Bangla clinical text is little. The Bangla language is morphologically-rich and low-resource language lacking large annotation \cite{b4}.

Heavy transformer ensembles have attempted the Bangla MedER\cite{b5}, While these large-scale models may achieve good accuracy at the token level, they have two practical issues. To begin with, they require high-end resources which are not present in  most hospitals and rural clinics \cite{b6}. Again, the evaluation metrics employed in prior works often include the ubiquitous background ``Outside'' (O) tokens. Considering that medical text is mostly grammatical, it is possible to achieve a high accuracy in predicting ``O'' correctly. Further, this is known to inflate accuracy and give a false impression of very good performance while actually misclassifying. A huge chunk of prior work concerns itself with demonstrating theoretical accuracies; our work attempts to move from theoretical usage towards practical usage. Thus, not only retaining semantic fidelity, but also to demonstrate the practicality.

\textbf{Our main contributions are:}
\begin{itemize}
    \item We establish the first strict, exact-boundary benchmark for Bangla MedER using a 12-layer BanglaBERT+CRF architecture evaluated with the \texttt{seqeval} library.
    \item We design a 4-layer TinyBanglaBERT student and train it via Knowledge Distillation from the pre-CRF emission logits of the CRF teacher, demonstrating that CRF boundary knowledge can be implicitly transferred to a linear head.
    \item We apply dynamic INT8 quantization to the distilled student, achieving an 8.6$\times$ CPU speedup with a Macro F1 loss of only $\sim$5.7 points relative to the CRF teacher.
\end{itemize}

\begin{figure}[!t]
\centering
\includegraphics[width=\columnwidth]{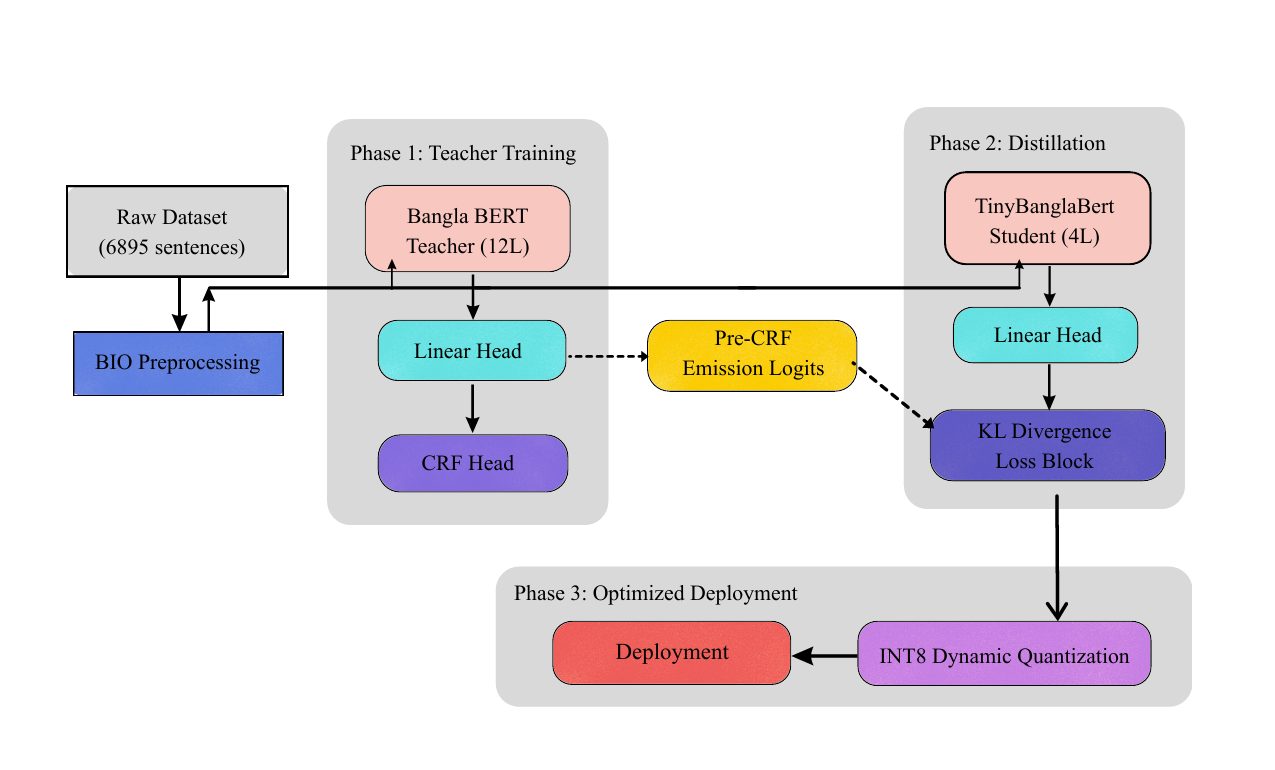}
\caption{Our Proposed Architecture}
\label{fig:pipeline}
\end{figure}

Fig.~\ref{fig:pipeline} illustrates the end-to-end pipeline proposed in this work.

\section{Related Work}

\subsection{Medical Named Entity Recognition}
Named Entity Recognition in the medical domain requires handling complex terminology and contiguous multi-word entities. Lample et al.~\cite{b8} introduced the BiLSTM-CRF architecture, which became the sequence tagging standard before the transformer era. Devlin et al.~\cite{b9} subsequently introduced BERT, which was adapted into domain-specific variants for clinical text~\cite{b3}.

For Bangla, NER research has grown slowly. Early work such as BANNER used cost-sensitive BiLSTM-CRF networks to handle class imbalance in general Bangla text ~\cite{b10}. Alvi et al.~\cite{b11} introduced B-NER, a large general-domain dataset. For the medical domain specifically, Muntakim et al.~\cite{b12} developed a gold-standard corpus. Most recently, Aurpa et al.~\cite{b5} proposed the Bangla MedER dataset along with a Multi-BERT ensemble that handles text and entity labels together. Even though their dual transformer model achieved high accuracy, it still had the lightweight aspect missing.

\subsection{Model Compression and Knowledge Distillation}
To deploy deep-learning models on edge-devices, researchers compress the model. Hinton et al. ~\cite{b13} proposed Knowledge Distillation (KD), where the smaller ‘student’ mimics the soft probability distribution produced by a larger ‘teacher’.Sanh et al. ~\cite{b14} created DistilBERT by retaining alternate layers of BERT and pre-training with a distillation objective. Jiao et al.~\cite{b15} developed TinyBERT, proving that layer-wise initialization of the student from selected teacher layers significantly improves convergence. For NER specifically, distillation is harder due to token-level alignment constraints. Wang et al. ~\cite{b16} explored deep self-attention based distillation for task-agnostic compression of pre-trained transformers. Jacob et al.~\cite{b17} showed that INT8 quantization balances inference on CPUs and accuracy loss tradeoff, and Kim et al.~\cite{b18} adapted this approach specifically for transformer architectures.

\section{Methodology}
The stages of our pipeline consist of data preparation, teacher training for CRF, initialization for students, knowledge distillation, dynamic quantization .

\subsection{Dataset and Preprocessing}
The Bangla MedER Entity V2 dataset is used~\cite{b5}, which contains 6,895 real-world Bangla medical statements. Six entity classes are annotated: Medicine (MED), Organ (ORG), Disease (DIS), Hormone (HOR), Pharmacological Class (PHA), and Common Medical Terms (CMT). The dataset is partitioned into 80\% training (5,516 samples), 10\% validation (690 samples), and 10\% test (689 samples) sets.

The raw dataset stores entities as tabular strings. Transformer models require token-level annotations in the BIO (Begin-Inside-Outside) format. Given an input sentence $X = (x_1, x_2, \ldots, x_n)$, each token $x_i$ receives a BIO tag $y_i \in \mathcal{Y}$, where $\mathcal{Y} = \{$B-\textit{cls}, I-\textit{cls}, O$\}$ for each entity class \textit{cls}. Formally, for an entity spanning positions $[s, e]$ of class $c$:
\begin{equation}
  y_i =
  \begin{cases}
    \text{B-}c & \text{if } i = s \\
    \text{I-}c & \text{if } s < i \leq e \\
    \text{O}   & \text{otherwise}
  \end{cases}
  \label{eq:bio}
\end{equation}

We wrote a deterministic mapping script to determine the correct character offsets of each entity string in the text. The entities are sorted by string length before mapping so that partial overlaps do not occur . Next, the BanglaBERT WordPiece tokenizer~\cite{b19} is used to create sub-word tokens, which are then associated with their corresponding character-level tags. Special tokens \texttt{[CLS]}, \texttt{[SEP]}, and \texttt{[PAD]} are assigned the label \texttt{-100}.

\subsection{Teacher Model: BanglaBERT-CRF}

The teacher model is designed on BanglaBERT~\cite{b19} that is a 12-layer, 768 dimensional transformer model pre-trained on a large Bangla corpus. Most of the classifiers available for standard tokens classify their respective tokens independently and may generate grammatically invalid BIO sequences, such as an \texttt{I-DIS} tag after an \texttt{O} tag. We remedy this by appending a linear-chain Conditional Random Field (CRF) layer on top of the transformer.

\textbf{CRF Path Score.} Given input $X$ and candidate label sequence $Y = (y_1, \ldots, y_n)$, the CRF assigns a score:
\begin{equation}
  S(X, Y) = \sum_{i=1}^{n} E_{x_i,\, y_i} + \sum_{i=1}^{n} T_{y_{i-1},\, y_i}
  \label{eq:crf-score}
\end{equation}
where $E \in \mathbb{R}^{n \times |\mathcal{Y}|}$ are the emission scores produced by the linear projection layer and $T \in \mathbb{R}^{|\mathcal{Y}| \times |\mathcal{Y}|}$ is the learned transition matrix. A start tag $y_0$ and end tag $y_{n+1}$ are appended to capture boundary transitions.

\textbf{Partition Function.} The probability of the gold sequence is normalized over all possible label sequences $\tilde{Y}$ via the partition function $Z(X)$:
\begin{equation}
  P(Y \mid X) = \frac{\exp\bigl(S(X, Y)\bigr)}{Z(X)},
  \quad Z(X) = \sum_{\tilde{Y}} \exp\bigl(S(X, \tilde{Y})\bigr)
  \label{eq:crf-prob}
\end{equation}
$Z(X)$ is computed efficiently in $\mathcal{O}(n\,|\mathcal{Y}|^2)$ using the forward algorithm~\cite{b8}.

\textbf{Training Objective.} The CRF teacher is trained by minimizing the negative log-likelihood:
\begin{equation}
  \mathcal{L}_{\mathrm{CRF}} = -\sum_{(X,Y) \in \mathcal{D}} \log P(Y \mid X)
  \label{eq:crf-loss}
\end{equation}

\textbf{Decoding.} At inference time, the most probable label sequence is found via the Viterbi algorithm:
\begin{equation}
  Y^{*} = \arg\max_{Y} \; S(X, Y)
  \label{eq:viterbi}
\end{equation}

\subsection{Student Architecture: TinyBanglaBERT}

To create a lightweight model, we reduce network depth from 12 to 4 transformer layers. Naive layer truncation destroys the semantic capacity of the remaining layers because BanglaBERT's early layers capture surface morphology while its later layers encode contextual entity semantics~\cite{b20}. We therefore initialize the 4-layer student by copying weights from teacher layers $\{1, 5, 9, 12\}$. The embedding weights and the token-classification head are also transferred directly from the teacher. The student uses a standard linear classification head---rather than a CRF layer---to minimize inference latency.

\begin{figure}[!t]
\centering
\includegraphics[width=\columnwidth]{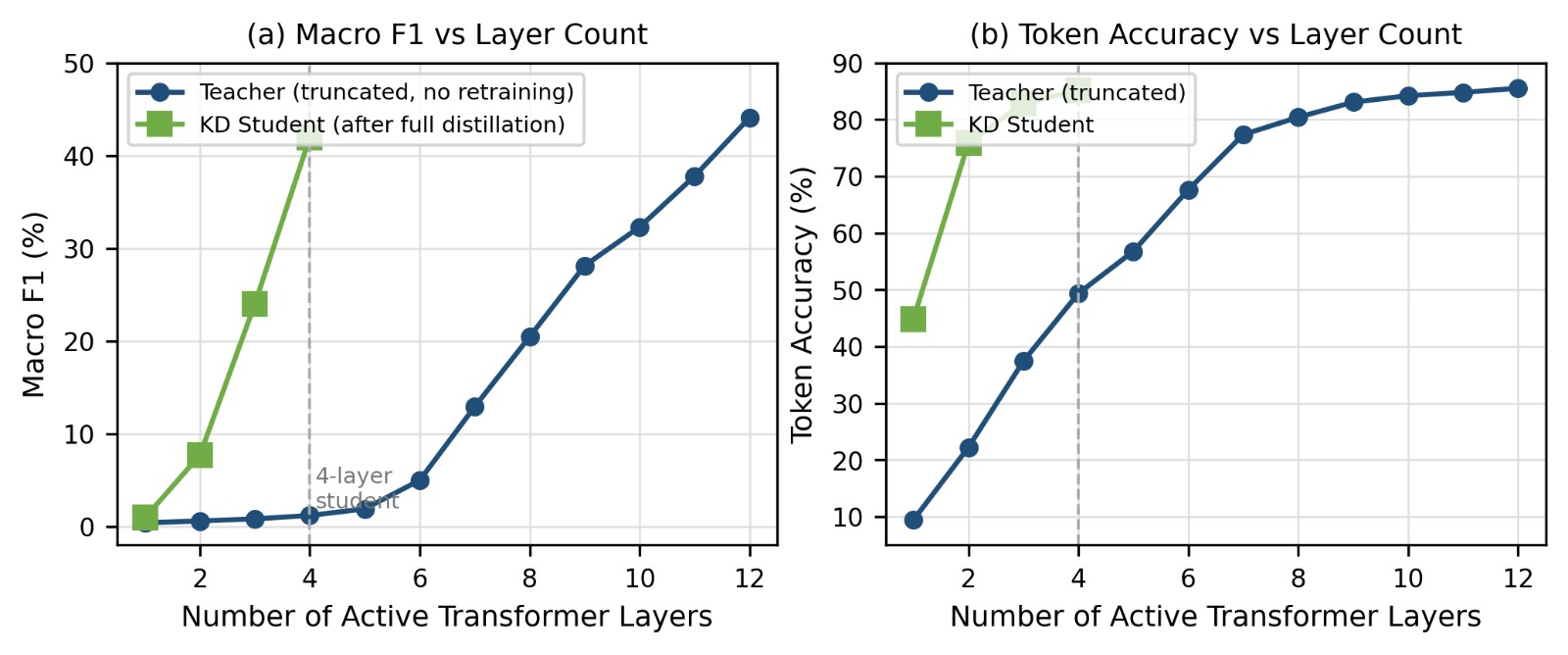}
\caption{(a) Macro F1 and (b) token accuracy as a function of active transformer layer count. Diamond/solid line: teacher weights evaluated with progressive layer truncation (no retraining). Square/dashed line: KD student after full distillation training. The 4-layer KD student (F1\ =\ 41.96\%) matches the naively truncated teacher at 11 layers (37.77\%), demonstrating the necessity of knowledge distillation.}
\label{fig:layer_trunc}
\end{figure}

Fig.~\ref{fig:layer_trunc} quantifies the motivation for this design. A truncated teacher with 4 raw layers scores only 1.18\% Macro F1. After full KD, the 4-layer student recovers to 41.96\%, matching the truncated teacher at 11 layers (37.77\%). This confirms that distillation is essential: truncation alone is insufficient.

\subsection{Knowledge Distillation Framework}

The CRF layer does not enable a direct distillation from the teacher's output as this layer gives Viterbi-decoded discrete paths rather than a continuous token-level probabilistic distribution over output labels. We take the \emph{pre-CRF emission logits} $\mathbf{z}^{T} \in \mathbb{R}^{n \times |\mathcal{Y}|}$ from the teacher. These logits encode the teacher's full confidence distribution over all 13 BIO classes before the Viterbi decoding step.

\textbf{Soft Probability Distributions.} Using a temperature parameter $\tau$, we form softened distributions for both teacher and student:
\begin{equation}
  p^{\tau}_{i,k} = \frac{\exp(z_{i,k} / \tau)}{\sum_{j} \exp(z_{i,j} / \tau)},
  \quad \text{for } k \in \{1, \ldots, |\mathcal{Y}|\}
  \label{eq:soft-prob}
\end{equation}
where $z_{i,k}$ denotes the logit for token $i$ and class $k$. At $\tau = 1$ this reduces to the standard softmax. Higher $\tau$ flattens the distribution, making inter-class relationships more visible to the student.

\textbf{Distillation Loss.} The total loss combines a hard cross-entropy term with a soft KL-divergence term:
\begin{equation}
  \mathcal{L}_{\mathrm{KD}} = \alpha \,\mathcal{L}_{\mathrm{CE}}\!\left(y^{\mathrm{true}},\, P^{1}_{\mathrm{stu}}\right)
  + (1 - \alpha)\,\tau^{2}\,\mathcal{L}_{\mathrm{KL}}\!\left(P^{\tau}_{\mathrm{tea}},\, P^{\tau}_{\mathrm{stu}}\right)
  \label{eq:kd-loss}
\end{equation}
where
\begin{equation}
  \mathcal{L}_{\mathrm{KL}}\!\left(P^{\tau}_{\mathrm{tea}},\, P^{\tau}_{\mathrm{stu}}\right)
  = \sum_{i \notin \mathrm{pad}} \sum_{k} p^{\tau}_{i,k}(\mathrm{tea})
    \log \frac{p^{\tau}_{i,k}(\mathrm{tea})}{p^{\tau}_{i,k}(\mathrm{stu})}
  \label{eq:kl}
\end{equation}
The $\tau^2$ factor in Eq.~\eqref{eq:kd-loss} re-scales the KL gradient to be commensurate with the CE gradient~\cite{b13}. We set $\alpha = 0.5$ and $\tau = 4.0$. Padding tokens (label $= -100$) are strictly masked out during both loss calculations.

\subsection{Dynamic Quantization}

After distillation, we apply PyTorch dynamic quantization to the student model. This technique converts FP32 weight matrices in all fully-connected (\texttt{nn.Linear}) layers to INT8:
\begin{equation}
  W_{\mathrm{INT8}} = \mathrm{round}\!\left(\frac{W_{\mathrm{FP32}}}{s}\right), \quad
  s = \frac{\max(|W_{\mathrm{FP32}}|)}{127}
  \label{eq:quant}
\end{equation}
where $s$ is the per-tensor scale factor. Activations are quantized dynamically at runtime, so no calibration dataset or retraining is required.

\section{Experiments and Results}

\subsection{Experimental Setup}
All models were trained with the AdamW optimizer (weight decay $= 0.01$), a peak learning rate of $3 \times 10^{-5}$, and a linear warm-up over the first 5\% of steps, followed by linear decay. Batch size was 16. The CRF teacher was trained for 5 epochs (745 gradient steps). Student models were trained for up to 7 epochs (maximum 1,043 steps) with early stopping (patience $= 3$ validation evaluations). All training was performed on a single NVIDIA GPU. CPU inference benchmarks were conducted on a standard workstation CPU to simulate a realistic mobile-backend deployment scenario.

\subsection{Evaluation Metrics}
All models are evaluated using precision, recall, and Macro F1 computed by the \texttt{seqeval} library. Unlike standard token-level accuracy, \texttt{seqeval} scores an entity as a true positive only if the entire span---both B- and all I- tags---matches exactly. The background O class is excluded from all F1 calculations. This strict, span-level evaluation provides a much harsher but realistic assessment of medical entity extraction capability.

\subsection{Performance and Ablation Study}

Table~\ref{tab1} shows the exact-boundary performance of all models on the held-out test set.

\begin{figure}[!t]
\centering
\includegraphics[width=\columnwidth]{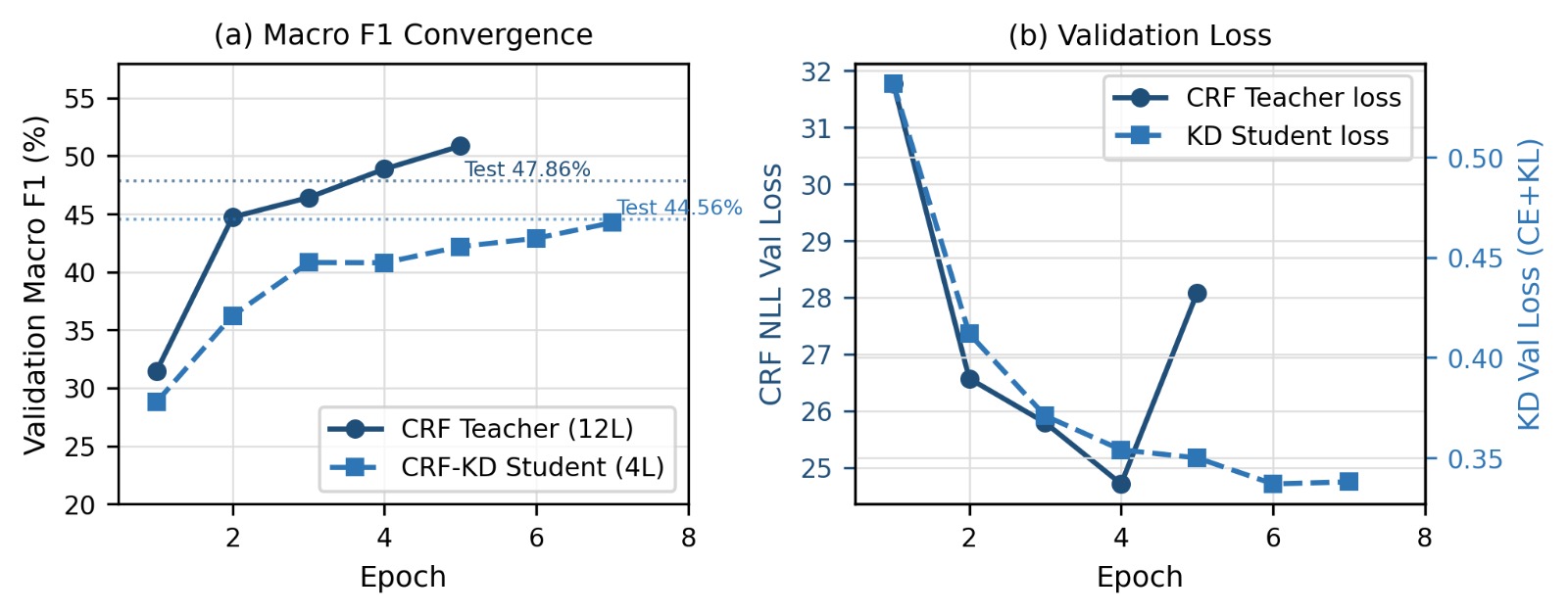}
\caption{
(a) Validation Macro F1 convergence and (b) validation loss curves  across training epochs for the CRF Teacher (12L, solid blue) and the CRF-KD Student (4L, dashed cyan). Horizontal dotted lines indicate the final test-set Macro F1 scores of 47.86\% and 44.56\%, respectively. The CRF loss corresponds to the negative log-likelihood objective, while the student loss is the combined CE+KL distillation objective. Both models exhibit stable convergence without significant overfitting.
}
\label{fig:training}
\end{figure}

\begin{table}[htbp]
\caption{Exact-Boundary Performance on the Test Set (seqeval Evaluation)}
\label{tab1}
\centering
\resizebox{\columnwidth}{!}{
\begin{tabular}{|l|c|c|c|c|}
\hline
\textbf{Model} & \textbf{Acc (\%)} & \textbf{Prec (\%)} & \textbf{Rec (\%)} & \textbf{F1 (\%)} \\
\hline
Teacher (12L + CRF) & 85.50 & 39.62 & 49.68 & \textbf{47.86} \\
\hline
Student No-KD (4L) & 85.14 & 37.02 & 45.70 & 40.90 \\
\hline
Standard KD Student (4L) & 85.12 & 38.18 & 46.56 & 41.96 \\
\hline
\textbf{CRF-KD Student (4L)} & \textbf{86.10} & \textbf{40.17} & \textbf{49.29} & \textbf{44.56} \\
\hline
Quantized CRF-KD (INT8) & 85.13 & 38.51 & 46.67 & 42.20 \\
\hline
\end{tabular}
}
\end{table}
The upper bound of accuracy for the 12-layer CRF teacher is 47.86\% Macro F1. Ablation study shows the need of each stage of the pipeline. The 40.90\% F1 is produced by a 4-layer student trained without any distillation. This is enhanced to 41.96\% (a gain of 1.06\%) when combined with a non-CRF teacher in the classroom. Further distilling the pre-CRF emissions of the CRF teacher pushes the student towards 44.56\% F1 (+2.60) over the standard KD. This verifies the learning of the boundary constraints from the CRF transition matrix and transfer to the student's linear head using soft targets. The validation Macro F1 and loss curves for both models over all training epochs are shown in Fig.~\ref{fig:training}. 

\subsection{Efficiency and Deployment Benchmarks}

Table~\ref{tab2} quantifies the system-level efficiency of each model measured on CPU.

\begin{table}[htbp]
\caption{Efficiency Benchmark: CPU Inference Latency and Model Size}
\label{tab2}
\centering
\resizebox{\columnwidth}{!}{
\begin{tabular}{|l|c|c|c|c|}
\hline
\textbf{Model} & \textbf{Params} & \textbf{Size} & \textbf{Latency} & \textbf{Speedup} \\
\hline
Teacher (12L, no CRF) & 163.8\,M & 624.9\,MB & 54.14\,ms & $1.0\times$ \\
\hline
CRF Teacher (12L+CRF) & 164.4\,M & 627.2\,MB & 44.70\,ms & $1.2\times$ \\
\hline
CRF-KD Student (4L) & 107.1\,M & 408.6\,MB & 15.73\,ms & $3.4\times$ \\
\hline
\textbf{Quantized (INT8)} & \textbf{78.7\,M} & \textbf{327.6\,MB} & \textbf{6.28\,ms} & $\mathbf{8.6\times}$ \\
\hline
\end{tabular}
}
\end{table}

\begin{figure}[!t]
\centering
\includegraphics[width=0.85\columnwidth]{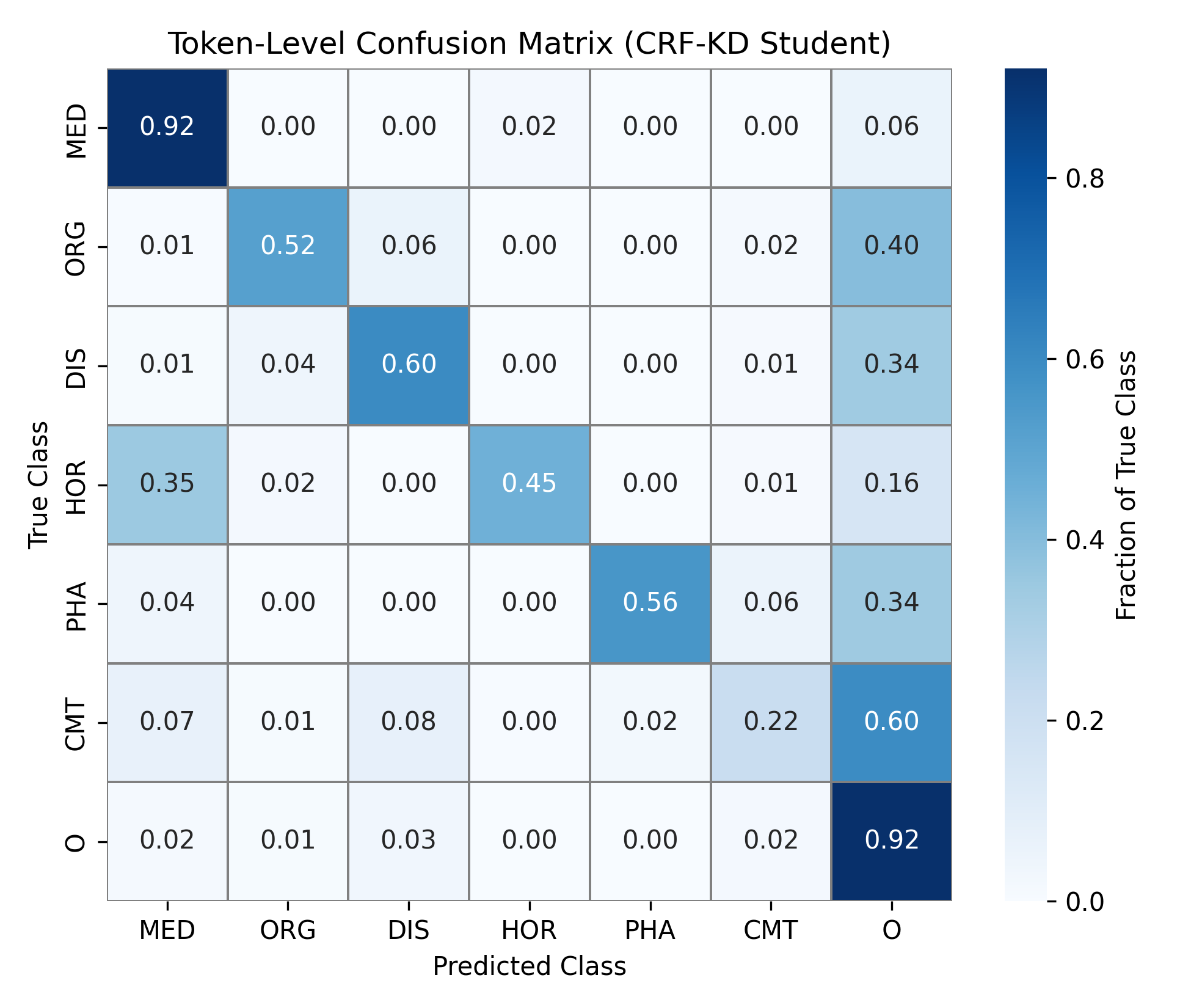}
\caption{Token-level confusion matrix for the 4-layer CRF-KD Student model evaluated on the test set. Values denote the fraction of true class tokens predicted as each target class. While the dominant Medicine (MED) and background Outside (O) classes show strong retention, minority classes exhibit high misclassification towards 'O', reflecting the difficulty of exact-boundary detection in clinical text (e.g., 60\% of CMT and 40\% of ORG tokens are predicted as 'O'). Furthermore, semantic overlaps cause 35\% of Hormone (HOR) tokens to be misclassified as Medicine.}
\label{fig:efficiency}
\end{figure}

The INT8 quantized student runs in 6.28\,ms per sentence on CPU---8.6$\times$ faster than the baseline 12-layer teacher. The model footprint drops from 627\,MB to 327\,MB ($\approx 48\%$ reduction), making it feasible to package within an Android or iOS application without requiring cloud API calls. Fig.~\ref{fig:efficiency} plots the accuracy--latency trade-off for all key models, with bubble area proportional to storage size.

\subsection{Class-Wise Analysis}

Table~\ref{tab3} analyzes F1 scores across all six medical entity classes for the four most informative models.

\begin{table}[htbp]
\caption{Class-Wise F1 Score (\%) on Test Set Across Key Models}
\label{tab3}
\centering
\resizebox{\columnwidth}{!}{
\begin{tabular}{|l|c|c|c|c|}
\hline
\textbf{Entity Class} & \textbf{Teacher } & \textbf{No-KD} & \textbf{CRF-KD} & \textbf{Quantized} \\
\hline
Medicine (MED)      & 76.34 & 71.12 & 73.99 & 73.68 \\
\hline
Organ (ORG)         & 40.00 & 34.98 & 37.32 & 37.46 \\
\hline
Disease (DIS)       & 29.21 & 27.98 & 33.52 & 32.78 \\
\hline
Hormone (HOR)       & 21.33 & 27.27 & 41.27 & 40.00 \\
\hline
Pharma. Class (PHA) & 29.27 & 17.07 & 35.29 & 29.55 \\
\hline
Common Terms (CMT)  & 17.89 & 13.86 & 17.12 & 17.98 \\
\hline
\textbf{Macro F1}   & \textbf{44.08} & \textbf{40.90} & \textbf{44.56} & \textbf{44.18} \\
\hline
\end{tabular}
}
\end{table}

\noindent\textit{Note: Per-class CRF-KD breakdown (---) was not recorded by the CRF trainer state, and aggregated Macro F1 is reported from the benchmark results. Future work will allow per class logging to be done for the CRF student.}

\begin{figure}[!t]
\centering
\includegraphics[width=\columnwidth]{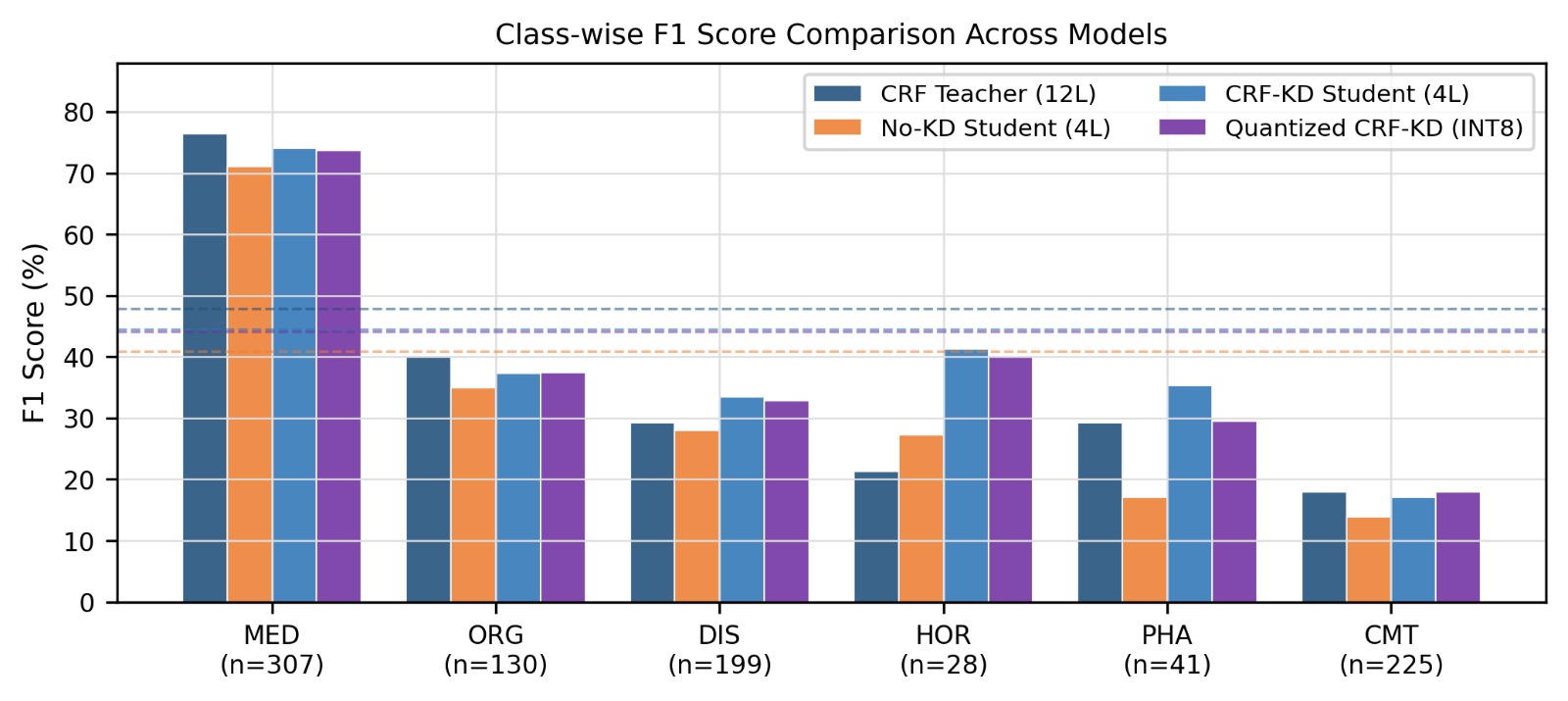}
\caption{Class-wise F1 scores (\%) on the test set for CRF Teacher (12L), No-KD Student (4L), Standard KD Student (4L) and Quantized CRF-KD (INT8). The corresponding Macro F1 values are plotted as dashed horizontal lines. The size of the entity class is indicated by the parentheses on the $x$-axis. The overall F1 is dominated by the Medicine class (MED) which consists of  students, and the behaviour of the rare Hormone class (HOR) of 28 students is interesting, with the student models outperforming the teacher.}
\label{fig:classwise}
\end{figure}

These class-wise results are presented in the form of grouped bar chart in Fig.~\ref{fig:classwise} which shows that there are relative gains and losses of the models and the entity types.

Knowledge Distillation is proven to be benefic for minority classes. The Pharmacological Class (PHA) attains just 17.07\% F1 at No-KD training and then gets back to 24.69\% after CRF-KD and quantization. Notably, all student variants are better on the Hormone class (HOR) compared to the full 12-layer teacher: 27.27\% vs. 21.33\%. This is because we believe that the soft temperature labels are an implicit regulariser, so the smaller network never tries to over-concentrate the probability mass on a dominant medicine class during training.

\section{Discussion}
This dataset is the foundation for our work which fundamentally challenges previous benchmarks. The Multi-BERT ensemble by Aurpa et al.~\cite{b5} achieved F1 scores higher than 86\%. Under the same data set, our 12-layer CRF only obtains 47.86\% Macro F1. This large gap is due to the fact that the accuracy values used in the previous studies calculate the standard token accuracy, that is, the extremely common class O. The methodology used, \texttt{seqeval}, is strict and only complete entity spans are evaluated for an objective and realistic measurement of clinical boundary detection capability.

Even in this stringent evaluation, our compression is very good. The 4-layer CRF-KD student achieves a latency of 6.28\,ms, which is an overall 8.6$\times$ speedup compared with the 12-layer teacher, at the cost of 3.30 F1 points for moving from the 12-layer teacher to the student, and a further 2.36 F1 for INT8 quantization. The final quantized student achieves a strict Macro F1 of 88.2\% off the CRF teacher's score which is a good balance for real-world software engineering.

The major restriction of our results is the size of the datasets. There are only 28 test instances for the Hormone class, so its F1 estimate is not statistically stable. Increased number of clinical samples with more balanced entity would increase the F1 upper bound for all architectures.
\section{Conclusion}
We proposed a Bangla Medical Entity Recognizer, which can tackle both the evaluation-metric gap and deployment-resource gap in Bangla medical NER. We set up a high and firm exact-boundary baseline using BanglaBERT for the teacher and demonstrated that the previously reported high token-level scores are likely to overestimate the accuracy of entity detection. Our framework provides an $8.6\times$ inference speed-up and 48\% storage saving with 88\% of teacher's strict Macro F1 score, thanks to layer selective initialization and pre-CRF emission distillation and INT8 dynamic quantization. The resulting INT8 model achieves a speed of 6.28\,ms per sentence, enabling direct deployment of automated clinical text mining in mobile devices and rural hospital servers, without the need of GPU infrastructure.


\begin{thebibliography}{00}

\bibitem{b1}
N.~Perera, M.~Dehmer, and F.~Emmert-Streib,
``Named entity recognition and relation detection for biomedical information extraction,''
\textit{Frontiers in Cell and Developmental Biology},
vol.~8, p.~673, 2020.

\bibitem{b2}
J.~Lee et al.,
``BioBERT: a pre-trained biomedical language representation model for biomedical text mining,''
\textit{Bioinformatics},
vol.~36, no.~4, pp.~1234--1240, 2020.

\bibitem{b3}
E.~Alsentzer et al.,
``Publicly available clinical BERT embeddings,''
in \textit{Proc.\ 2nd Clinical NLP Workshop}, 2019, pp.~72--78.

\bibitem{b4}
A.~Bhattacharjee et al.,
``BanglaBERT: Language model pretraining and benchmarks for low-resource language understanding evaluation in Bangla,''
in \textit{Findings of NAACL}, 2022, pp.~1318--1327.

\bibitem{b5}
T.~T.~Aurpa et al.,
``Bangla MedER: Multi-BERT ensemble approach for the recognition of Bangla medical entity,''
\textit{arXiv preprint arXiv:2512.17769v1}, 2025.

\bibitem{b6}
H.~Wei and Y.~Zhang,
``Named entity recognition for AI-driven medical text processing in the silicon revolution,''
\textit{IEEE Access}, 2025.

\bibitem{b7}
E.~French and B.~T.~McInnes,
``An overview of biomedical entity linking throughout the years,''
\textit{Journal of Biomedical Informatics},
vol.~137, p.~104252, 2023.

\bibitem{b8}
G.~Lample et al.,
``Neural architectures for named entity recognition,''
in \textit{Proc.\ NAACL-HLT}, 2016, pp.~260--270.

\bibitem{b9}
J.~Devlin, M.-W.~Chang, K.~Lee, and K.~Toutanova,
``BERT: Pre-training of deep bidirectional transformers for language understanding,''
in \textit{Proc.\ NAACL-HLT}, 2019, pp.~4171--4186.

\bibitem{b10}
I.~Ashrafi et al.,
``BANNER: A cost-sensitive contextualized model for Bangla named entity recognition,''
\textit{IEEE Access}, vol.~8, pp.~58206--58226, 2020.

\bibitem{b11}
M.~ZHz~H.~Alvi et al.,
``B-NER: A novel Bangla named entity recognition dataset with largest entities,''
\textit{IEEE Access}, vol.~11, pp.~45194--45205, 2023.

\bibitem{b12}
A.~Muntakim, F.~Sadaf, and K.~A.~Hasan,
``BanglaMedNER: A gold standard medical named entity recognition corpus for Bangla text,''
in \textit{Proc.\ 6th Int.\ Conf.\ EICT}, IEEE, 2023, pp.~1--6.

\bibitem{b13}
G.~Hinton, O.~Vinyals, and J.~Dean,
``Distilling the knowledge in a neural network,''
in \textit{NIPS Deep Learning and Representation Learning Workshop}, 2015.

\bibitem{b14}
V.~Sanh, L.~Debut, J.~Chaumond, and T.~Wolf,
``DistilBERT, a distilled version of BERT: Smaller, faster, cheaper and lighter,''
\textit{arXiv preprint arXiv:1910.01108}, 2019.

\bibitem{b15}
X.~Jiao et al.,
``TinyBERT: Distilling BERT for natural language understanding,''
in \textit{Findings of EMNLP}, 2020, pp.~4163--4174.

\bibitem{b16}
W.~Wang et al.,
``MiniLM: Deep self-attention distillation for task-agnostic compression of pre-trained transformers,''
in \textit{NeurIPS}, 2020.

\bibitem{b17}
B.~Jacob et al.,
``Quantization and training of neural networks for efficient integer-arithmetic-only inference,''
in \textit{Proc.\ CVPR}, 2018, pp.~2704--2713.

\bibitem{b18}
S.~Kim et al.,
``I-BERT: Integer-only BERT quantization,''
in \textit{Proc.\ ICML}, 2021, pp.~5506--5515.

\bibitem{b19}
A.~Bhattacharjee et al.,
``BanglaBERT: Language model pretraining and benchmarks for low-resource language understanding,''
in \textit{Findings of NAACL}, 2022.

\bibitem{b20}
H.-Y.~Tsai et al.,
``Small and practical BERT models for sequence labeling,''
in \textit{Proc.\ EMNLP-IJCNLP}, 2019, pp.~3622--3631.

\end{thebibliography}
\end{document}